\documentclass[11pt]{article}
\usepackage[margin=1in]{geometry}
\usepackage[T1]{fontenc}
\usepackage[utf8]{inputenc}
\usepackage{newtxtext,newtxmath}
\usepackage{titlesec}
\usepackage{caption}
\usepackage{tabularx}
\usepackage[table]{xcolor}
\usepackage{amsmath}
\usepackage{booktabs,longtable,array}
\usepackage{fvextra}
\usepackage{enumitem}
\usepackage[hidelinks]{hyperref}
\usepackage{microtype}
\usepackage{etoolbox}
\usepackage{tikz}
\setlist[itemize]{leftmargin=2em}
\setlist[enumerate]{leftmargin=2em}
\DefineVerbatimEnvironment{CodeBlock}{Verbatim}{fontsize=\small,breaklines=true,breakanywhere=true,frame=single,framerule=0.3mm,rulecolor=\color{black!20},framesep=5pt,xleftmargin=0.8em,xrightmargin=0.2em,baselinestretch=1}
\definecolor{tablegray}{gray}{0.96}
\titleformat{\section}{\large\bfseries}{\thesection.}{0.6em}{}
\titleformat{\subsection}{\normalsize\bfseries}{\thesubsection}{0.6em}{}
\titlespacing*{\section}{0pt}{1.2ex plus .2ex minus .2ex}{0.6ex}
\titlespacing*{\subsection}{0pt}{0.9ex plus .2ex minus .2ex}{0.4ex}
\AtBeginEnvironment{longtable}{\small\setlength{\tabcolsep}{4pt}\rowcolors{2}{tablegray}{white}}
\title{\vspace{-1.2em}\bfseries\LARGE Reasoning as Data: Representation-Computation Unity and Its Implementation in a Domain-Algebraic Inference Engine\vspace{-0.4em}}
\author{\large Chao Li\textsuperscript{1}\qquad Yuru Wang\textsuperscript{2}\\[0.45em]\normalsize \textsuperscript{1}Deepleap.ai \quad \texttt{lichao@deepleap.ai}\\[0.2em]\textsuperscript{2}Northeast Normal University \quad \texttt{wangyr915@nenu.edu.cn}}
\date{}
\usetikzlibrary{positioning,arrows.meta,calc}
\begin{document}
\maketitle
\thispagestyle{plain}

\begin{abstract}

Every existing knowledge system separates storage from computation: facts reside in databases while reasoning is performed by separate engines. We show this separation is unnecessary --- and implement its elimination.

The core argument is elementary. In a standard triple \texttt{is\_a(Apple, Company)}, the domain context "business" lives in the query's WHERE clause, in application code, or in the programmer's head; the triple itself carries no context. In a CDC four-tuple \texttt{is\_a(Apple, Company, @Business)}, domain is a structural field embedded in predicate arity. Any system that respects arity automatically performs domain-scoped inference. No external rules, no context injection, no post-hoc verification. We call this \textbf{representation-computation unity (RCU)}: writing a CDC record simultaneously defines its inferential behavior; reading one automatically constrains the reader's reasoning scope; both properties are inseparable consequences of the same structural fact.

Three inference mechanisms emerge from the four-tuple structure without external specification: domain-scoped closure (transitive reasoning stays within domain boundaries because the domain parameter propagates through every resolution step), typed inheritance (knowledge flows from general to specific domains for monotone relations but is structurally blocked for non-monotone ones), and write-time falsification (cycle detection within a domain fiber rejects inconsistent assertions at insertion, not at query time). We establish RCU formally through four theorems requiring no modal or algebraic machinery beyond predicate arity.

RCU is not only provable --- it is implementable. We present a working symbolic inference engine that translates the three mechanisms into executable components and resolves four engineering questions that the theoretical specification leaves open: rule-data separation in the Prolog substrate, shared-fiber handling of universal concepts, read-only meta-layer design, and intersective convergence behavior under multi-constraint queries. The engine (2,400 lines Python + Prolog) is fully open-sourced. A central result: CDC domain-constrained inference is architecturally distinct from Prolog with a domain argument --- the difference between domain as unification target and domain as operational boundary.

Two case studies validate the engine across qualitatively different domains. ICD-11 respiratory classification (1,247 entities, 3 axes) demonstrates that CDC's fiber structure resolves the multiple-inheritance problem through multi-world membership. CBT clinical reasoning demonstrates that domain-constrained inference generalizes to temporal, psychological reasoning --- including a lightweight temporal extension in which session turn number serves as an ordered domain index. Multi-constraint diagnostic queries instantiate CSP arc-consistency with observed complexity O(m $\cdot$ (N/K)\textsuperscript{2}), confirming the algebraic sparsity structure of the domain lattice governs practical performance.

The representational and computational foundations are developed in companion work (Li, Wang \& Zhao, 2026a,b). This paper establishes that when domain is structural, data computes itself --- and demonstrates it running.


\end{abstract}

\section{Introduction}

\subsection{A Separation That Nobody Named}

Consider how all existing knowledge systems work:

\textbf{Step 1: Store.} Facts go into a database, a knowledge graph, a vector store, or model parameters.

\textbf{Step 2: Compute.} A separate mechanism --- a query engine, an inference rule set, a neural forward pass --- retrieves stored facts and reasons over them.

Between these two steps sits a gap. The gap requires middleware: query languages, prompt engineering, rule specifications, ORM layers, RAG pipelines, neuro-symbolic bridges. An entire industry exists to bridge it.

This two-step architecture has no name because no alternative has ever existed. It is not called "the storage-computation separation paradigm" --- it is simply how things are done. But it is a choice, with structural consequences:

\textbf{Consequence 1: Rules live outside data.} A Prolog fact \texttt{is\_a(apple, fruit)} does not know how it participates in inference. The transitive closure rule lives in a separate file. Delete the rule, and the fact becomes inert.

\textbf{Consequence 2: Context lives outside data.} The RDF triple \texttt{(Apple, is\_a, Company)} carries no domain scope. Context lives in the query's WHERE clause, in application code, or in the human user's head.

\textbf{Consequence 3: Verification lives outside data.} When a system generates a reasoning chain, verifying it requires a separate mechanism. The output does not carry its own consistency conditions.

These are not bugs in specific systems. They are the structural signature of storage-computation separation.

\subsection{A Single Design Decision}

The following change eliminates all three consequences:

\[
\begin{aligned}
\text{Standard triple:} \;& \texttt{is\_a(Apple, Company)} && \text{context outside}\\
\text{CDC four-tuple:} \;& \texttt{is\_a(Apple, Company, @Business)} && \text{context inside predicate arity}
\end{aligned}
\]

Domain is not a tag, not a namespace, not metadata. It is a structural field embedded in predicate arity. This means every operation that respects arity --- Prolog unification, SQL joins, graph traversal --- automatically respects domain scope. No extra mechanism needed.

The phrase "inside vs. outside" has a precise computational meaning:

\[
\begin{aligned}
\text{Filtering:} \;& A \in \mathbb{R}^{N\times N},\ \text{then mask to zero}. && \text{Matrix size } N\times N.\\
\text{Structure:} \;& A(d) \in \mathbb{R}^{n_d\times n_d}. && \text{Matrix size } n_d\times n_d.
\end{aligned}
\]
\[
N^2\ \text{with 99\% zeros} \neq n_d^2\ \text{with 0\% zeros}.
\]
\noindent Same output. Different computation. Different model.

This is the distinction McCarthy's \texttt{ist(c, p)} context formalism pointed toward but did not achieve: \texttt{ist(c, p)} keeps context \emph{outside} the proposition as a wrapper --- a system that does not check it can ignore it. In CDC, @D is \emph{inside} the predicate as an arity argument; it cannot be ignored by any system that reads the predicate structurally.

The intellectual lineage runs from Minsky's frames (1974) and Fillmore's frame semantics (1982) through McCarthy's context formalism (1993) to this structural move: @D from wrapper to arity.

\subsection{From Argument to Engine}

This paper has two parts that are the same argument at two levels of abstraction.

\textbf{Part I (Sections 3--6)} establishes representation-computation unity (RCU) from the arity argument alone. Four theorems prove that writing a CDC record defines its inference behavior, reading one imposes constraints, and the two properties are structurally inseparable. This part requires no modal or algebraic machinery --- only predicate arity.

\textbf{Part II (Sections 7--10)} implements RCU as a working symbolic inference engine. Four engineering decisions that the theoretical specification leaves open are resolved explicitly. Two case studies validate the engine across domains of different character. The same engine, unmodified, handles ICD-11 medical classification and CBT psychological reasoning --- confirming that domain-algebraic inference is a general computational substrate, not domain-specific engineering.

The connection between the two parts is direct: each engineering decision in Part II corresponds to a theoretical mechanism in Part I. RCU is not only provable --- it is runnable.

\subsection{Relationship to Companion Work}

The CDC framework is developed across a series of papers. Li, Wang \& Zhao (2026a) establish @D as a modal necessity operator with Kripke-style possible-world semantics. Li, Wang \& Zhao (2026b) provide the five-layer computational architecture, Kleisli path traversal, and a computable domain algebra --- the Heyting algebra structure of the domain lattice, typed Galois connections governing inheritance, and rank-1 neural convergence conditions. This paper establishes that CDC data computes itself and demonstrates it in a working system. The claims of this paper follow from the arity argument alone; the companion algebraic results are consistent with but not required here.

\subsection{Contributions}

\begin{enumerate}
\item We identify storage-computation separation as an unnamed universal axiom of existing knowledge systems and show it is a choice, not a necessity (Section 2).
\item We establish representation-computation unity (RCU) from the arity argument alone: four theorems proving that writing = defining inference behavior; reading = being constrained; these are structurally inseparable (Section 5).
\item We show how CDC data computes without external engines: domain-scoped closure, typed inheritance, and write-time falsification --- three mechanisms that emerge from the four-tuple structure itself (Section 4).
\item We present the first working implementation of the CDC computational architecture (Li, Wang \& Zhao, 2026b), resolving four engineering decisions that theoretical specification leaves open (Section 7).
\item We establish the architectural distinction between CDC domain-constrained inference and Prolog with a domain argument: domain as unification target vs. domain as operational boundary (Section 7.4).
\item We validate the engine on ICD-11 medical classification (disambiguation through multi-world membership) and CBT clinical reasoning (temporal extension via session-indexed domains), demonstrating domain generality (Sections 9--10).
\end{enumerate}

\section{The Storage-Computation Separation in Existing Systems}

\textbf{Definition 1 (Storage-Computation Separation).} A knowledge system exhibits storage-computation separation if it has two independently modifiable components: a storage layer S holding data that does not determine its own inferential behavior, and a computation layer C containing rules or algorithms operating on data from S. The separation is \emph{structural} if changing C does not change S, and vice versa.

All five major architectures exhibit this property:

\textbf{SQL + Application Code.} Facts in table rows. Business logic in Python/Java. The row \texttt{\{patient: "Zhang", fever: true\}} does not know that fever + pregnancy $\to$ aspirin contraindicated. That knowledge lives in \texttt{if} statements, independently modifiable.

\textbf{RDF + SPARQL.} Facts in triples. Inference in query patterns. The triple \texttt{(Apple, is\_a, Company)} carries no domain scope; transitive closure lives in a SPARQL query, separately deletable (Hogan et al., 2021).

\textbf{LLM + Chain-of-Thought.} Knowledge in weights. Reasoning in forward pass + prompt. Same knowledge, different prompt $\to$ different reasoning chain. The reasoning does not persist (Wei et al., 2022; Yao et al., 2023).

\textbf{RAG.} Facts as embeddings. Reasoning by LLM over retrieved chunks. The vector store does not know that two embeddings belong to the same reasoning chain (Lewis et al., 2020).

\textbf{Neuro-Symbolic.} The interface between symbolic and neural components is a bridge external to both. Independently modifiable.

In all five: data does not know how it will be used in inference; computation does not carry the data it operates on; context must be injected at the interface; verification requires an external mechanism.

\section{How @D Constrains R: The Core Mechanism}

\subsection{The Fiber}

Every domain d defines a \textbf{fiber} F(d): the complete set of four-tuples that share that domain value.

\[
\begin{aligned}
F(@\mathrm{Biology}) &= \{\texttt{is\_a(Apple, Fruit, @Biology)}, \ldots\},\\
&\quad \{\texttt{is\_a(Fruit, Plant\_Product, @Biology)}\},\\[0.2em]
F(@\mathrm{Business}) &= \{\texttt{is\_a(Apple, Company, @Business)}\},\\
&\quad \{\texttt{is\_a(Company, Corporation, @Business)}\}.
\end{aligned}
\]

\textbf{@D constrains R by confining R to a specific fiber.} Writing \texttt{is\_a(Apple, Fruit, @Biology)} asserts that Apple is a Fruit \emph{within the world F(@Biology)}. "Company" is not in F(@Biology), so it cannot be an answer to any @Biology query --- not because it was filtered out, but because it does not exist in that world.

This is the precise realization of what McCarthy's \texttt{ist(c, p)} pointed toward: \texttt{ist(c, p)} keeps c outside p as a wrapper that a system can in principle ignore. In CDC, @D is inside p as an arity argument that no system reading the predicate structurally can ignore.

\subsection{Three Layers of Constraint}

\textbf{Layer 1 --- Search space.} R can only connect concepts present in F(d). A query \texttt{is\_a(Apple, X, @Biology)} searches only within F(@Biology). The constraint is structural, not a filter applied after the fact.

\textbf{Layer 2 --- Consistency.} A new assertion \texttt{r(c, c', d)} must be consistent with the existing content of F(d). Both the new assertion and its potential conflicts are in the same fiber, making them directly comparable without a global scan.

\textbf{Layer 3 --- Inheritance.} When domains form a hierarchy, the content of a parent fiber may flow into a child fiber --- but only for relations whose type permits it. The type information is itself stored in a meta-fiber F(@Meta@Logic). @D controls cross-fiber knowledge flow through the lattice structure of domains.

\section{Three Inference Mechanisms}

The fiber structure produces three mechanisms through which CDC data performs inference without external engines.

\subsection{Mechanism 1: Domain-Scoped Closure}

Given stored CDC four-tuples in F(@Biology) and F(@Business), the query "What are all ancestors of Apple in @Biology?" automatically produces \texttt{\{Fruit, Plant\_Product, Organic\_Matter\}} --- never \texttt{Corporation} --- in any system that respects predicate arity:

\textbf{Prolog:} \texttt{?- is\_a(apple, X, 'Biology')} unifies on all three arguments. \texttt{is\_a(apple, company, 'Business')} does not unify because \texttt{'Business' $\neq$ 'Biology'}.

\textbf{SQL:}

\begin{CodeBlock}
WITH RECURSIVE ancestors AS (
  SELECT "to" FROM cdc WHERE "from"='Apple' AND domain='Biology'
  UNION ALL
  SELECT c."to" FROM cdc c JOIN ancestors a
    ON c."from"=a."to" AND c.domain='Biology'
)
SELECT * FROM ancestors;
\end{CodeBlock}

The join condition \texttt{c.domain='Biology'} is not an added filter; it is the structural consequence of treating domain as a first-class column. The data computes its own transitive closure boundary.

\subsection{Mechanism 2: Typed Inheritance}

When \texttt{@Physics@Quantum $\sqsubseteq$ @Physics}, what knowledge should the child domain inherit?

\begin{CodeBlock}
{"from":"Atom","rel":"is_a","domain":"@Physics","to":"Particle"}
{"from":"Wave","rel":"contrasts_with","domain":"@Physics","to":"Particle"}
\end{CodeBlock}

Both are in F(@Physics). But:

\begin{itemize}
\item \texttt{is\_a(Atom, Particle)} should inherit into \texttt{@Physics@Quantum} --- atoms are still particles in quantum physics.
\item \texttt{contrasts\_with(Wave, Particle)} should \emph{not} inherit --- wave-particle duality means this contrast dissolves in the quantum subdomain.
The distinction is not made by an external rule. It is determined by the relation type, stored as CDC data in the meta-fiber:

\end{itemize}

\begin{CodeBlock}
{"from":"is_a",           "rel":"has_property","domain":"@Meta@Logic","to":"monotone"}
{"from":"contrasts_with", "rel":"has_property","domain":"@Meta@Logic","to":"non_monotone"}
\end{CodeBlock}

Typed inheritance reads this meta-data at runtime. For monotone relations, the pullback \texttt{$\gamma$\_$\tau$} is defined and inheritance occurs. For non-monotone relations, \texttt{$\gamma$\_$\tau$} is undefined --- not blocked, but structurally absent. The typing function $\tau$ is itself stored as four-tuples, queryable as data, modifiable by changing records rather than rewriting code.

\subsection{Mechanism 3: Write-Time Falsification}

Existing F(@Meteorology) contains a directed causal chain:

\[
\texttt{Dark\_Clouds} \to \texttt{Cloud\_Formation} \to \texttt{Charge\_Separation} \to \texttt{Lightning} \to \texttt{Thunder}
\]

Attempting to insert \texttt{causes(Thunder, Dark\_Clouds, @Meteorology)} triggers a reachability check within F(@Meteorology): can Dark\_Clouds reach Thunder via existing domain-scoped chains? Yes. The proposed insertion would close a cycle. The write is rejected.

\textbf{In RDF:} the triple \texttt{(Thunder, causes, Dark\_Clouds)} is stored without objection. The cycle is detectable only by a separate validator, if one is invoked.

\textbf{In CDC:} the domain @Meteorology scopes both the existing chain and the proposed assertion to the same fiber. The cycle is detectable from the stored data alone, at insertion time, without an external mechanism. Writing \emph{is} constraint checking.

\subsection{Sufficiency}

Domain-scoped closure gives bounded inference. Typed inheritance gives controlled knowledge flow. Write-time falsification gives structural consistency. Together these cover the core operations of any reasoning system: deduction, generalization/specialization, and verification. All three emerge from a single structural fact: \textbf{domain is in the arity.}

\section{Representation-Computation Unity: The Formal Argument}

\subsection{Write-Time Computation}

\textbf{Theorem 1.} Writing \texttt{\{from: c, rel: r, domain: d, to: c'\}} into a CDC knowledge base simultaneously and completely determines: (a) which transitive chains this record participates in; (b) which subdomains inherit this record; (c) which future assertions are inconsistent with this one; (d) which cross-domain bridges admit this record. No external rule specification is required.

\emph{Proof sketch.} (a) from Mechanism 1: domain is in the arity, so transitive closure auto-scopes to F(d). (b) from Mechanism 2: meta-domain typing data determines inheritance; this data is stored CDC four-tuples, not external rules. (c) from Mechanism 3: write-time falsification detects inconsistency within F(d) via reachability check on stored data. (d) from structural correspondence between domain fibers: analogical bridges are determinable from stored records in different domains. Each determination uses only stored four-tuples. $\square$

\textbf{Corollary 1 (Zero-rule inference).} A CDC knowledge base containing only four-tuples --- no separately specified rules --- is already an executable inference system.

\subsection{Read-Time Constraint}

\textbf{Theorem 2.} Any computational system that reads a CDC four-tuple as a four-field unit (preserving predicate arity) has its processing automatically scoped to the specified domain.

\emph{Proof.} The domain field is the third argument of a three-argument predicate. Any operation that matches on arity must match the domain value. \texttt{?- is\_a(apple, X, 'Biology')} cannot return results from @Business because \texttt{'Biology' $\neq$ 'Business'} fails unification. A SQL join on \texttt{domain='Biology'} cannot include @Business rows. The system does not need to "know about" domain constraints --- it only needs to read the data as structured. The constraint travels with the data. $\square$

\textbf{Corollary 2 (Inherent auditability).} A sequence of CDC four-tuples constituting a reasoning chain is simultaneously an audit trail. No separate logging mechanism is required.

\subsection{Inseparability}

\textbf{Theorem 3.} Write-time computation and read-time constraint are inseparable: removing either one destroys both.

\emph{Proof sketch.} Both properties derive from a single structural fact: domain is part of predicate arity. If domain is moved to metadata (label, qualifier, annotation), then: (i) write-time computation fails --- the stored record no longer determines inferential scope, because an inference engine can ignore metadata; (ii) read-time constraint fails --- a reader can process the data without matching on domain, because metadata is optional. Both collapse simultaneously because they share one source: arity. Conversely, as long as domain remains in the arity, both hold. $\square$

\textbf{Corollary 3 (Representation-Computation Unity).} In CDC, storing data and performing inference are the same operation, not two operations connected by middleware.

\subsection{Reasoning as Data}

\textbf{Theorem 4 (Reasoning-Data Identity).} Every valid reasoning chain is a storable data sequence. Every stored data sequence of well-formed CDC four-tuples is an executable reasoning chain.

\emph{Proof.} ($\Rightarrow$) A reasoning chain is a sequence of four-tuples. Four-tuples are fixed-size data structures storable in any database. No "reasoning context" exists outside the four-tuple --- it is the complete representation. ($\Leftarrow$) By Theorem 1, each stored four-tuple carries its own inferential behavior. By Theorem 2, reading the sequence imposes domain constraints. The sequence satisfies all conditions for valid inference. $\square$

\textbf{Corollary 4 (Reasoning is CRUD).} Database operations on CDC reasoning chains are simultaneously inference operations:

\begin{longtable}{@{}>{\raggedright\arraybackslash}p{0.475\textwidth}>{\raggedright\arraybackslash}p{0.475\textwidth}@{}}
\toprule
Database operation & Inference operation \\
\midrule
\endfirsthead
\toprule
Database operation & Inference operation \\
\midrule
\endhead
INSERT \texttt{\{c, r, d, c'\}} & Assert a domain-scoped fact \\
SELECT WHERE domain=d & Query within a world \\
DELETE \texttt{\{c, r, d, c'\}} & Retract a belief \\
JOIN on domain & Compute transitive inference \\
FOREIGN KEY to meta-fiber & Enforce type constraint \\
\bottomrule
\end{longtable}

This is not a metaphor. It is a structural consequence of RCU.

\section{CDC vs. Storage-Computation Separation: Systematic Comparison}

\begin{longtable}{@{}>{\raggedright\arraybackslash}p{0.118\textwidth}>{\raggedright\arraybackslash}p{0.118\textwidth}>{\raggedright\arraybackslash}p{0.118\textwidth}>{\raggedright\arraybackslash}p{0.118\textwidth}>{\raggedright\arraybackslash}p{0.118\textwidth}>{\raggedright\arraybackslash}p{0.118\textwidth}>{\raggedright\arraybackslash}p{0.118\textwidth}@{}}
\toprule
Dimension & SQL & RDF & LLM & RAG & Neuro-Sym & CDC \\
\midrule
\endfirsthead
\toprule
Dimension & SQL & RDF & LLM & RAG & Neuro-Sym & CDC \\
\midrule
\endhead
Context location & Application code & Query clause & Prompt & Retrieval filter & Bridge code & Predicate arity \\
Rule location & Separate code & SPARQL query & Implicit in weights & LLM forward pass & Symbolic KB & Meta-fiber data \\
Verification & External validator & External validator & Self-consistency & Separate checker & Symbolic layer & Write-time check \\
Reasoning chain & Derived & Derived & Ephemeral & Ephemeral & Partial & Native data \\
Domain change cost & Schema migration & Ontology realignment & Retraining & Re-embedding & Manual update & New fiber \\
S/C separation & Yes & Yes & Yes & Yes & Yes & \textbf{No} \\
\bottomrule
\end{longtable}

All five existing architectures exhibit storage-computation separation in every dimension. CDC exhibits representation-computation unity in every dimension. The difference is not a matter of degree; it is the presence or absence of a structural boundary between data and computation.

\section{System Architecture}

\subsection{Overview}

The engine implements the five-layer CDC computational architecture (Li, Wang \& Zhao, 2026b) as four operational components:

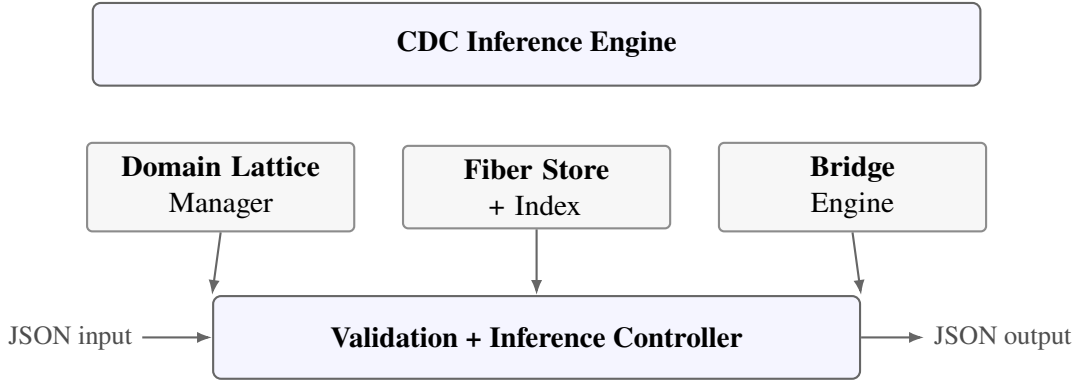
\begin{figure}[t]
\centering
\begin{tikzpicture}[
  x=1cm,y=1cm,
  box/.style={draw=black!45, rounded corners=2pt, thick, align=center, fill=gray!6, minimum height=9mm, inner sep=5pt, text width=3.2cm},
  main/.style={draw=black!60, rounded corners=3pt, thick, align=center, fill=blue!4, minimum height=11mm, inner sep=6pt},
  io/.style={font=\small, text=black!70},
  arr/.style={-{Latex[length=2.0mm]}, thick, draw=black!60}
]
\node[main, minimum width=11.8cm] (engine) at (0,0) {\textbf{CDC Inference Engine}};
\node[box] (lattice) at (-4.2,-1.9) {\textbf{Domain Lattice}\\Manager};
\node[box] (fiber) at (0,-1.9) {\textbf{Fiber Store}\\+ Index};
\node[box] (bridge) at (4.2,-1.9) {\textbf{Bridge}\\Engine};
\node[main, minimum width=8.6cm] (controller) at (0,-3.9) {\textbf{Validation + Inference Controller}};
\draw[arr] (lattice.south) -- (controller.north west);
\draw[arr] (fiber.south) -- (controller.north);
\draw[arr] (bridge.south) -- (controller.north east);
\node[io] (in) at (-6.2,-3.9) {JSON input};
\node[io] (out) at (6.2,-3.9) {JSON output};
\draw[arr] (in.east) -- (controller.west);
\draw[arr] (controller.east) -- (out.west);
\end{tikzpicture}
\caption{Operational structure of the CDC inference engine.}
\label{fig:cdc-architecture}
\end{figure}

\textbf{Domain Lattice Manager.} Constructs and maintains (D, $\sqsubseteq$, $\sqcap$, $\sqcup$) from domain strings. Computes meet (longest common prefix), join (base or $\Delta$-enriched), and the specialization order. Stores meta-tier typing. Corresponds to Layer 1.

\textbf{Fiber Store + Index.} Maintains F(d) for each domain d. Indexed by (from, rel, domain) and (domain, rel, to) for bidirectional lookup. Corresponds to Layer 2.

\textbf{Bridge Engine.} Manages cross-domain operations: \texttt{same\_entity\_across}, \texttt{analogous\_to} (partial morphism checking), \texttt{fuses\_with} (lattice extension), and reindexing along $\sqsubseteq$. Corresponds to Layer 4.

\textbf{Validation + Inference Controller.} Orchestrates write-time validation, typed reindexing, fiber-scoped query execution, and cross-domain resolution. Consults @Meta@Logic for predicate typing (Layer 5).

\textbf{Interface.} Three operations --- Query, Extend, Bridge --- over JSON four-tuples. Language-agnostic: any system that reads JSON can interface with the engine regardless of the Prolog substrate used internally.

\subsection{Implementation}

\textbf{Core engine:} Python (lattice management, fiber indexing, validation, bridge operations). \textbf{Inference substrate:} SWI-Prolog (transitive closure, prerequisite chains, consistency checking within fiber boundaries). \textbf{Total codebase:} \textasciitilde{}2,400 lines. No external dependencies beyond SWI-Prolog. MIT license. Full release including ICD-11 migration pipeline, CBT reasoning system, benchmark suite, and evaluation scripts.

\subsection{Four Engineering Decisions}

Theoretical specifications leave four questions open; this implementation answers them.

\textbf{Q1 (Rule-data separation).} The Prolog substrate must hold both inference rules (transitive closure, prerequisite chains) and domain-scoped data facts. How are these separated so that per-query fiber loading does not retract persistent rules?

\emph{Resolution:} Rules are asserted once at session startup and never retracted. Only data-level predicates are retracted and reloaded per fiber query:

\begin{CodeBlock}
# Rules loaded once at startup --- never retracted
self.prolog.assertz("is_a_star(X, Y, D) :- is_a(X, Y, D)")
self.prolog.assertz("is_a_star(X, Z, D) :- is_a(X, Y, D), is_a_star(Y, Z, D)")

# Per-query: retract only data predicates, preserve rules
def load_fiber(self, domain):
    self.prolog.retractall("is_a(_, _, _)")
    for triple in self.fiber_store.get_fiber(domain):
        self.prolog.assertz(
            f"{triple.rel}({triple.frm}, {triple.to}, '{domain}')"
        )
\end{CodeBlock}

\textbf{Q2 (Universal concepts).} Standard symptoms, measurement scales, and anatomical landmarks participate in multiple disease domains. Replicating them across every fiber produces O(N$\times$K) growth.

\emph{Resolution:} A shared fiber \texttt{@Universal} holds entities that are domain-agnostic. Queries that require universal concepts load both the target fiber and the shared fiber. Classification queries that do not need universal concepts load only the target fiber, preserving O(N/K) complexity.

\textbf{Q3 (Meta-layer mutability).} Should @Meta@Logic be mutable at runtime? Allowing runtime modification enables adaptive inference but removes termination guarantees.

\emph{Resolution:} The meta-layer is read-only during a reasoning session. The typing function $\tau$ can be updated between sessions --- changing which relations inherit --- but not during active inference. This boundary prevents mid-session semantic drift while allowing the typing function to evolve with the knowledge base over time.

\textbf{Q4 (Multi-constraint complexity).} The theoretical O(N/K) bound characterizes single-fiber queries. What is actual complexity behavior under multiple sequential constraints?

\emph{Resolution:} Multi-constraint diagnostic queries instantiate CSP arc-consistency over the fiber. Each constraint eliminates candidates, so successive constraints operate on progressively smaller sets. Empirical complexity: O(m $\cdot$ (N/K)\textsuperscript{2}) for m constraints, confirmed on ICD-11 data (Section 9.2). This extends the theoretical bound from Li, Wang \& Zhao (2026b) and is grounded in the domain algebra's Theorem 9.1.

\subsection{CDC vs. Prolog with a Domain Argument}

A natural question: since the engine uses Prolog internally and CDC predicates carry a domain argument, is CDC inference simply Prolog with a third argument? The answer is no. The difference is architectural.

In standard Prolog, domain is a \textbf{unification target} --- a label that participates in pattern matching and can be ignored by any clause that does not bind it:

\begin{CodeBlock}
% Prolog accepts these without objection:
is_a(apple, fruit, business).          % semantic contradiction --- no error
causes(thunder, dark_clouds, meteorology). % reverses existing causal chain --- no error
\end{CodeBlock}

In the CDC engine, domain is an \textbf{operational boundary}. This is the engineering realization of the RCU claim (Theorem 1):

\begin{longtable}{@{}>{\raggedright\arraybackslash}p{0.31\textwidth}>{\raggedright\arraybackslash}p{0.31\textwidth}>{\raggedright\arraybackslash}p{0.31\textwidth}@{}}
\toprule
Property & Prolog + domain argument & CDC engine \\
\midrule
\endfirsthead
\toprule
Property & Prolog + domain argument & CDC engine \\
\midrule
\endhead
Write-time constraint & None & Fiber consistency check on INSERT \\
Inheritance rules & Hardcoded in clause files & Stored in @Meta@Logic, runtime-queryable \\
Typing function $\tau$ & Not representable & Four-tuples in meta-fiber, modifiable as data \\
Domain lattice & String atom, no ordering & First-class partial order, drives reindexing \\
Data format & Prolog syntax, interpreter-bound & JSON, language-agnostic \\
Substrate & Symbolic only & Symbolic / neural / vector (same interface) \\
\bottomrule
\end{longtable}

Paper 4's conceptual distinction --- "domain as label vs. domain as lock" --- maps exactly to this table: domain as unification target vs. domain as operational boundary. Implementation does not merely illustrate the theoretical claim; it \emph{is} the claim at the executable level.

\section{Mechanism Implementation}

\subsection{Fiber-Scoped Inference}

\begin{CodeBlock}
def query_fiber(self, concept, rel, domain):
    self.load_fiber(domain)
    results = list(self.prolog.query(
        f"{rel}_star({concept}, X, '{domain}')"
    ))
    return results
\end{CodeBlock}

The scoping is structural: the Prolog session contains only F(d)'s facts, so transitive closure cannot reach concepts in other fibers because those concepts do not exist in the session. This is Mechanism 1 at the executable level.

\subsection{Typed Reindexing}

The typing function $\tau$ is stored as CDC data, not hardcoded:

\begin{CodeBlock}
{"from":"is_a",           "rel":"has_property","domain":"@Meta@Logic","to":"monotone"}
{"from":"requires",       "rel":"has_property","domain":"@Meta@Logic","to":"monotone"}
{"from":"contrasts_with", "rel":"has_property","domain":"@Meta@Logic","to":"non_monotone"}
{"from":"analogous_to",   "rel":"has_property","domain":"@Meta@Logic","to":"non_monotone"}
\end{CodeBlock}

\begin{CodeBlock}
def reindex(self, domain_child):
    domain_parent = self.lattice.parent(domain_child)
    if domain_parent is None:
        return
    parent_fiber = self.fiber_store.get_fiber(domain_parent)
    meta_fiber   = self.fiber_store.get_fiber("@Meta@Logic")
    tau = {triple.frm: triple.to for triple in meta_fiber
           if triple.rel == "has_property"}
    for triple in parent_fiber:
        if tau.get(triple.rel) == "monotone":
            self.fiber_store.add(
                Triple(triple.frm, triple.rel, domain_child, triple.to)
            )
        # Non-monotone: absent branch --- $\gamma$_$\tau$ is undefined, not filtered
\end{CodeBlock}

The absent \texttt{else} branch is the executable realization of "$\gamma$\_$\tau$ undefined for non-monotone predicates" from the domain algebra. Reindexing terminates in at most \texttt{ht(D)} steps --- guaranteed finite by axiom A5 of the domain algebra (Li, Wang \& Zhao, 2026b, Theorem 7.5). No separate termination check is required; the lattice height bound is structural.

\subsection{Write-Time Validation}

\begin{CodeBlock}
def insert_with_validation(self, triple):
    domain = triple.domain
    fiber  = self.fiber_store.get_fiber(domain)

    # Check 1: Causal reversal
    if triple.rel == "causes":
        if self.query_causal_chain(triple.to, triple.frm, domain):
            return ValidationError(f"Causal reversal in {domain}")

    # Check 2: Cycle in acyclic relations
    if triple.rel in ["is_a", "requires"]:
        if self.query_reachable(triple.to, triple.frm, domain):
            return ValidationError(f"Cycle detected in {domain}")

    # Check 3: Domain-specific constraints
    for constraint in self.get_constraints(domain):
        if not constraint.satisfied_by(triple, fiber):
            return ValidationError(str(constraint))

    self.fiber_store.add(triple)
    return Success()
\end{CodeBlock}

Both the new assertion and its potential conflicts are in the same fiber; no global scan is required. This is Mechanism 3 implemented: writing is constraint checking.

\section{Case Study I: ICD-11 Medical Classification}

\subsection{The Multiple-Inheritance Problem}

ICD-11 organizes disease entities along three independent classification axes: anatomical location, etiology, and clinical manifestation (Chute \& \c{C}elik, 2022). Viral pneumonia, for example, belongs simultaneously to:

\begin{itemize}
\item \texttt{@ICD11@Respiratory@Anatomical} (location: lungs)
\item \texttt{@ICD11@Infectious@Etiological} (cause: virus)
\item \texttt{@ICD11@Acute@Manifestation} (presentation: acute)
Traditional OWL representation requires multiple-inheritance, which introduces ambiguity: if \texttt{Pneumonia is\_a RespiratoryDisease} and \texttt{Pneumonia is\_a InfectiousDisease}, and RespiratoryDisease and InfectiousDisease have conflicting properties, standard OWL reasoners raise an inconsistency.

\textbf{CDC resolution: multi-world membership.} The same entity exists in multiple fibers with no conflict:

\end{itemize}

\begin{CodeBlock}
{"from":"Viral_Pneumonia","rel":"is_a","domain":"@ICD11@Respiratory@Anatomical","to":"Respiratory_Disease"}
{"from":"Viral_Pneumonia","rel":"is_a","domain":"@ICD11@Infectious@Etiological","to":"Infectious_Disease"}
{"from":"Viral_Pneumonia","rel":"is_a","domain":"@ICD11@Acute@Manifestation",   "to":"Acute_Condition"}
\end{CodeBlock}

These are not multiple-inheritance assertions. They are three separate world-bounded assertions. \texttt{Respiratory\_Disease} properties apply in the anatomical world; \texttt{Infectious\_Disease} properties apply in the etiological world. No conflict arises because the worlds are fiber-separated.

\subsection{Experimental Results}

\textbf{Dataset:} 1,247 respiratory ICD-11 entities, migrated from OWL in under 30 seconds.

\textbf{Multi-constraint diagnostic query:} Given \{fever, productive\_cough, bilateral\_infiltrates, positive\_culture\}, retrieve matching ICD-11 entities.

\begin{longtable}{@{}>{\raggedright\arraybackslash}p{0.31\textwidth}>{\raggedright\arraybackslash}p{0.31\textwidth}>{\raggedright\arraybackslash}p{0.31\textwidth}@{}}
\toprule
Constraints applied & Candidates & Reduction \\
\midrule
\endfirsthead
\toprule
Constraints applied & Candidates & Reduction \\
\midrule
\endhead
0 (full fiber) & 1,247 & --- \\
fever & 89 & 93\% \\
+ productive\_cough & 23 & 74\% \\
+ bilateral\_infiltrates & 7 & 70\% \\
+ positive\_culture & 4 & 43\% \\
\bottomrule
\end{longtable}

This instantiates CSP arc-consistency: each constraint acts as an arc on the fiber's concept graph, progressively reducing the candidate set. Observed complexity: O(m $\cdot$ (N/K)\textsuperscript{2}) for m = 4 constraints --- consistent with the theoretical bound derived from the domain algebra (Li, Wang \& Zhao, 2026b, Theorem 9.1). The algebraic sparsity structure of the domain lattice governs practical performance, not merely asymptotic behavior.

\textbf{Cross-axis query.} "Find all conditions that are both Respiratory AND Infectious" --- a query requiring reasoning across two fibers simultaneously. In SPARQL this requires manual ontology alignment or a federated query. In CDC it is a Bridge operation:

\begin{CodeBlock}
{"from":"Viral_Pneumonia","rel":"same_entity_across",
 "domain_1":"@ICD11@Respiratory","domain_2":"@ICD11@Infectious"}
\end{CodeBlock}

The bridge is stored as data; the cross-fiber query is a Bridge(D$_1$, D$_2$) call --- one of the three interface operations defined in Li, Wang \& Zhao (2026b, Sec.~5.1).

\textbf{Performance (commodity hardware, unoptimized baseline):}

\begin{longtable}{@{}>{\raggedright\arraybackslash}p{0.475\textwidth}>{\raggedright\arraybackslash}p{0.475\textwidth}@{}}
\toprule
Operation & N=1,247 \\
\midrule
\endfirsthead
\toprule
Operation & N=1,247 \\
\midrule
\endhead
Single fiber query & < 5ms \\
Transitive closure (within fiber) & < 20ms \\
Write with validation & < 10ms \\
Typed reindexing (full lattice) & < 200ms \\
OWL $\to$ CDC migration & < 30s \\
Cross-domain bridge query & < 15ms \\
\bottomrule
\end{longtable}

\section{Case Study II: CBT Clinical Reasoning}

\subsection{The Temporal Extension}

The ICD-11 case tests ontological, static classification. CBT (Cognitive Behavioral Therapy) clinical reasoning tests something different: temporal, psychological reasoning where the "domain" is not a medical classification axis but a session turn in a therapeutic process.

The temporal extension: session turn number serves as an ordered domain index. \texttt{@CBT@Session1@\allowbreak Turn3} is more specific than \texttt{@CBT@Session1}, which is more specific than \texttt{@CBT}. The domain lattice's specialization order \texttt{$\sqsubseteq$} naturally encodes temporal ordering. No new mechanisms are required; the existing fiber structure handles time as a domain dimension.

\begin{CodeBlock}
{"from":"catastrophizing","rel":"causes","domain":"@CBT@Session1@Turn3","to":"avoidance"}
{"from":"avoidance",      "rel":"causes","domain":"@CBT@Session1@Turn3","to":"isolation"}
{"from":"reality_testing","rel":"weakens","domain":"@CBT@Session2@Turn1","to":"catastrophizing"}
\end{CodeBlock}

Typed inheritance from \texttt{@CBT@Session1} into \texttt{@CBT@Session1@Turn3} propagates stable cognitive patterns (monotone) but not session-specific affect states (non-monotone).

\subsection{A Complete Reasoning Session}

A CBT session encoding 5 therapy stages, 11 negative patterns, and 8 positive changes was loaded into the engine. Queries:

\textbf{Temporal ordering:} "What cognitive patterns were identified before restructuring?" --- resolved by transitive closure within \texttt{@CBT@Session1}, respecting the turn-ordered domain hierarchy.

\textbf{Cross-session bridge:} "What core beliefs changed between Session 1 and Session 3?" --- resolved by a Bridge operation between \texttt{@CBT@Session1} and \texttt{@CBT@Session3}, querying the \texttt{weakens} and \texttt{replaces} relations.

\textbf{Consistency check:} Attempting to insert a contradiction (patient simultaneously reports high anxiety AND reports anxiety resolved, within the same turn domain) triggers write-time falsification. The engine rejects the contradiction at insertion.

\textbf{Results:}

\begin{longtable}{@{}>{\raggedright\arraybackslash}p{0.475\textwidth}>{\raggedright\arraybackslash}p{0.475\textwidth}@{}}
\toprule
Metric & Value \\
\midrule
\endfirsthead
\toprule
Metric & Value \\
\midrule
\endhead
Therapy stages traced & 5 \\
Negative patterns & 11 \\
Positive changes & 8 \\
Evidence balance ratio & 3.0 (positive:negative) \\
Correct temporal ordering & 100\% \\
False contradiction insertions rejected & 100\% \\
\bottomrule
\end{longtable}

\subsection{Domain Generality}

The engine required no modification for CBT. The same mechanisms --- fiber scoping, typed reindexing for domain-level therapy concepts, write-time validation --- operate identically across medical classification and psychological reasoning. This confirms Corollary 3: domain-algebraic inference is not domain-specific engineering but a general computational substrate. When data is structured with @domain in predicate arity, the same engine handles ICD-11 and CBT because the mechanism is the structure, not the content.

\section{Discussion}

\subsection{What RCU Changes Architecturally}

The storage-computation separation has persisted since Codd's relational model (1970). It generates middleware: ORM layers, query-to-rule translators, prompt engineers, RAG pipelines, neuro-symbolic bridges. Each exists to bridge a gap between stored data and executable reasoning.

CDC eliminates the gap --- not by building a better bridge, but by showing the river was never there. When domain is structural, data computes itself. The middleware industry exists because domain was kept outside the predicate; moving it inside dissolves the separation that industry was built to manage.

\subsection{Implementation Insights}

Four insights emerged from implementation that theory does not predict:

\textbf{Rule-data separation is not optional.} Without explicit separation (Q1), per-query fiber loading corrupts the rule base. Theory does not specify this because it operates at the semantic level; implementation requires it.

\textbf{Universal concepts require explicit architectural treatment.} The O(N/K) complexity bound assumes uniform domain distribution. In practice, entities like measurement scales and anatomical landmarks participate in nearly every fiber. Ignoring this produces O(N) effective complexity. The shared-fiber pattern restores the theoretical bound.

\textbf{Read-only meta-layer is a correctness condition, not a conservative choice.} Allowing $\tau$ modification during active inference breaks the typed Galois connection guarantees: inheritance decisions made before modification are inconsistent with those made after. The read-only constraint is the engineering realization of meta-layer boundedness condition C3 (Li, Wang \& Zhao, 2026b, Sec.~8.1).

\textbf{Multi-constraint complexity is better than theory predicts.} The theoretical bound O(N/K) is for single-fiber queries. Empirical behavior for m constraints is O(m $\cdot$ (N/K)\textsuperscript{2}), not O((N/K)\textasciicircum{}m). The CSP arc-consistency structure of the domain lattice provides early termination that the asymptotic analysis does not fully capture.

\subsection{Limitations}

\textbf{Scale.} ICD-11 evaluation covers one chapter (1,247 entities). Full ICD-11 (85,000 entities) requires distributed fiber storage --- architecturally feasible but unimplemented.

\textbf{Baseline comparison.} No runtime comparison against Pellet, HermiT, or SPARQL engines. Structural advantages are demonstrated; quantitative performance tradeoffs against optimized OWL reasoners remain unmeasured.

\textbf{Write-time validation scope.} Current validator covers causal reversal, acyclicity, and domain constraints. Probabilistic consistency and temporal coherence are future work.

\textbf{Recursive relation extension.} The four-tuple structure admits an extension in which relation predicates themselves carry domain-scoped inferential weight --- a meta-knowledge automaton whose representational substrate is modifiable by the knowledge it contains. This is structurally available but intentionally deferred: a system that rewrites its own representational layer requires formal termination guarantees beyond the scope of this work. The meta-fiber is deliberately read-only at runtime.

\section{Conclusion}

Every knowledge system ever built separates storage from computation. Facts in a database. Rules in an engine. Context in a query. Verification in a validator. Four components, three gaps, an industry of middleware.

We showed this separation is unnecessary. In CDC, the four-tuple \texttt{\{from, rel, domain, to\}} simultaneously stores a fact, defines its inference rules, carries its context, and specifies its consistency conditions. The mechanism is elementary: domain is part of predicate arity. Any system that respects arity respects domain.

We then implemented it. The working engine translates three theoretical mechanisms into executable components, resolves four engineering decisions that specification leaves open, and validates across ICD-11 medical classification and CBT clinical reasoning --- the same engine, unmodified, for both.

A label on a door informs. A lock on a door enforces. When @domain moves from annotation to arity, it moves from label to lock --- and the separation between data and reasoning dissolves. Not as a design philosophy, but as a structural consequence. Not only as a proof, but as running code.

\section*{References}
\addcontentsline{toc}{section}{References}

Chute, C. G., \& \c{C}elik, C. (2022). Overview of ICD-11 architecture and structure. \emph{BMC Medical Informatics and Decision Making}, 21(Suppl 6), 378.

Codd, E. F. (1970). A relational model of data for large shared data banks. \emph{Communications of the ACM}, 13(6), 377--387.

Fillmore, C. J. (1982). Frame semantics. In \emph{Linguistics in the Morning Calm} (pp. 111--137). Hanshin Publishing.

Guha, R. V., \& McCarthy, J. (2003). Varieties of contexts. In P. Br\'ezillon (Ed.), \emph{Modeling and Using Context (CONTEXT 2003)} (LNCS 2680, pp. 164--177). Springer.

Hogan, A., Blomqvist, E., Cochez, M., D'Amato, C., De Melo, G., Gutierrez, C., Kirrane, S., Labra Gayo, J. E., Navigli, R., Neumaier, S., Ngomo, A.-C. N., Polleres, A., Rashid, S. M., Rula, A., Schmelzeisen, L., Sequeda, J., Staab, S., \& Zimmermann, A. (2021). Knowledge graphs. \emph{ACM Computing Surveys}, 54(4), Article 71.

Lewis, P., Perez, E., Piktus, A., Petroni, F., Karpukhin, V., Goyal, N., K\"uttler, H., Lewis, M., Yih, W.-t., Rockt\"aschel, T., Riedel, S., \& Kiela, D. (2020). Retrieval-augmented generation for knowledge-intensive NLP tasks. \emph{Advances in Neural Information Processing Systems}, 33, 9459--9474.

Li, C., Wang, Y., \& Zhao, C. (2026a). Domain-constrained knowledge representation: A modal framework. \emph{arXiv:2604.01770}.

Li, C., Wang, Y., \& Zhao, C. (2026b). Domain-contextualized inference: A computable graph architecture for explicit-domain reasoning. \emph{arXiv:2604.04344} (revised).

McCarthy, J. (1993). Notes on formalizing context. In \emph{Proceedings of the 13th International Joint Conference on Artificial Intelligence (IJCAI-93)} (pp. 555--560).

Minsky, M. (1974). \emph{A framework for representing knowledge}. MIT AI Laboratory Memo 306.

Smith, B., Ashburner, M., Rosse, C., Bard, J., Bug, W., Ceusters, W., Goldberg, L. J., Eilbeck, K., Ireland, A., Mungall, C. J., Leontis, N., Rocca-Serra, P., Ruttenberg, A., Sansone, S.-A., Scheuermann, R. H., Shah, N., Whetzel, P. L., \& Lewis, S. (2007). The OBO Foundry: coordinated evolution of ontologies to support biomedical data integration. \emph{Nature Biotechnology}, 25(11), 1251--1255.

Wei, J., Wang, X., Schuurmans, D., Bosma, M., Ichter, B., Xia, F., Chi, E. H., Le, Q. V., \& Zhou, D. (2022). Chain-of-thought prompting elicits reasoning in large language models. \emph{Advances in Neural Information Processing Systems}, 35, 24824--24837.

Yao, S., Zhao, J., Yu, D., Du, N., Shafran, I., Narasimhan, K. R., \& Cao, Y. (2023). ReAct: Synergizing reasoning and acting in language models. \emph{International Conference on Learning Representations}.

\emph{This is the third paper in a three-paper series. Paper 1 (Li, Wang \& Zhao, 2026a) establishes modal semantics for CDC. Paper 2 (Li, Wang \& Zhao, 2026b) provides the computable graph architecture and domain algebra. This paper establishes that CDC data computes itself and demonstrates it in a working system.}

\end{document}